\documentclass[12pt]{article}

\usepackage[utf8]{inputenc} 
\usepackage[T1]{fontenc}    
\usepackage{hyperref}       
\usepackage{url}            
\usepackage{booktabs}       
\usepackage{amsfonts}       
\usepackage{nicefrac}       
\usepackage{microtype}      
\usepackage{xcolor}         
\usepackage{algorithm}
\usepackage{algpseudocode}
\usepackage{graphicx}
\usepackage{float} 
\usepackage{booktabs}
\usepackage{soul}
\usepackage{tabularx}
\usepackage{amsmath}
\usepackage{caption}
\usepackage{amsthm}
\usepackage{cite}
\usepackage[numbers,sort&compress]{natbib}  
\usepackage{geometry}
\usepackage{times} 
\usepackage{setspace}
\usepackage{adjustbox}
\usepackage{graphicx} 
\usepackage{pifont} 
\usepackage{array} 
\usepackage{booktabs}
\usepackage{lipsum}
\usepackage{tikz}
\usepackage{makecell} 
\newcommand{\cmark}{\ding{51}} 
\newcommand{\xmark}{\ding{55}}
\usepackage{amsmath, amssymb, algorithm, algorithmicx, algpseudocode, graphicx}
\usepackage{algorithm}
\usepackage{algpseudocode}
\usepackage{amsmath}
\usepackage{graphicx}
\usepackage{varwidth}
\usepackage{tikz}
\usetikzlibrary{shapes.geometric, arrows.meta, positioning, shadows, backgrounds}
\usepackage{helvet}
\date{}
\theoremstyle{plain} 

\theoremstyle{definition} 

\title{\Large \textbf{KANGURA}: \textbf{\underline{K}}olmogorov-\textbf{\underline{A}}rnold \textbf{\underline{N}}etwork-Based \textbf{\underline{G}}eometry-Aware Learning with \textbf{\underline{U}}nified \textbf{\underline{R}}epresentation \textbf{\underline{A}}ttention for 3D Modeling of Complex Structures}

\usepackage[affil-it]{authblk} 

\usepackage[affil-it]{authblk}

\makeatletter
\newcommand{\repeatthanks}{\textsuperscript{\@fnsymbol{1}}}
\makeatother

\author[1]{\textbf{Mohammad Reza Shafie}\thanks{These authors are joint first authors and contributed equally to this work.}}
\author[2]{\textbf{Morteza Hajiabadi}\repeatthanks}
\author[3, 4]{\textbf{Hamed Khosravi}\repeatthanks}
\author[5]{\textbf{Mobina Noori}}
\author[3]{\textbf{Imtiaz Ahmed}\thanks{Corresponding author: imtiaz.ahmed@mail.wvu.edu}}

\affil[1]{Department of Electrical Engineering, Iran University of Science and Technology, Tehran, Iran}
\affil[2]{Department of Computer Engineering, Iran University of Science and Technology, Tehran, Iran}
\affil[3]{Department of Industrial \& Management Systems Engineering, West Virginia University, Morgantown, WV, USA}
\affil[4]{H. Milton Stewart School of Industrial and Systems Engineering, 
Georgia Institute of Technology, Atlanta, GA, USA}
\affil[5]{Department of Computer Science, University of California, Davis, Davis, CA, USA}

\begin{document}

\maketitle
\begin{abstract}
Microbial Fuel Cells (MFCs) offer a promising pathway for sustainable energy generation by converting organic matter into electricity through microbial processes. A key factor influencing MFC performance is the anode structure, where design and material properties play a crucial role. Existing predictive models struggle to capture the complex geometric dependencies necessary to optimize these structures. To solve this problem, we propose \textbf{KANGURA}: \textbf{\underline{K}}olmogorov-\textbf{\underline{A}}rnold \textbf{\underline{N}}etwork-Based \textbf{\underline{G}}eometry-Aware Learning with \textbf{\underline{U}}nified \textbf{\underline{R}}epresentation \textbf{\underline{A}}ttention. KANGURA introduces a new approach to three-dimensional (3D) machine learning modeling. It formulates prediction as a function decomposition problem, where Kolmogorov-Arnold Network (KAN)-based representation learning reconstructs geometric relationships without a conventional multi-layer perceptron (MLP). To refine spatial understanding, geometry-disentangled representation learning separates structural variations into interpretable components, while unified attention mechanisms dynamically enhance critical geometric regions. Experimental results demonstrate that KANGURA outperforms over 15 state-of-the-art (SOTA) models on the ModelNet40 benchmark dataset, achieving 92.7\% accuracy, and excels in a real-world MFC anode structure problem with 97\% accuracy. This establishes KANGURA as a robust framework for 3D geometric modeling, unlocking new possibilities for optimizing complex structures in advanced manufacturing and quality-driven engineering applications.
\end{abstract}

\section{Introduction}
Microbial Fuel Cells (MFCs) have gained increasing attention as a promising sustainable energy technology, utilizing microorganisms to convert organic matter into electricity. This dual-purpose system not only generates renewable energy but also aids in wastewater treatment and organic waste management \citep{zhang2024preparation}. A key factor influencing MFC efficiency is the anode structure, which plays a crucial role in electron transfer from microorganisms to the circuit. Optimizing anode design can significantly enhance power generation and improve substrate degradation rates \citep{nguyen2020comparative}. When effectively implemented at scale, MFC technology offers a broad range of applications, including electricity generation, biohydrogen production, wastewater treatment, biosensing, hazardous waste remediation, and integration into robotic systems \citep{Chakma2025}.

Three-dimensional (3D) models enable detailed simulation of fluid flow and ion transport within complex anode geometries by solving governing equations over volumetric meshes. This allows the identification of localized performance issues, which are critical for optimizing MFC anode design.
However, their high computational cost limits their practicality for iterative structural prediction \cite{kesler2023comparing}. 
By incorporating kinetics and electrochemical equations such as Monod and Butler–Volmer formulations, 3D models capture spatial heterogeneities in substrate utilization and electron transfer. Despite the computational demands, they remain essential tools for guiding the rational design of high-performance MFC architectures \cite{JADHAV2021124256}.

Despite its potential, predicting the optimal 3D structure of MFC anodes remains a challenge. The interplay between geometric complexity, material constraints, and microbial interactions makes it difficult to develop models that accurately capture these relationships. Many traditional approaches struggle to balance the intricate trade-offs between anode shape, material composition, and biofilm formation, often resulting in suboptimal designs \citep{oyedeji2023optimal}. Furthermore, the performance of anodes is highly dependent on material properties, as variations can directly impact biofilm growth and electron transfer efficiency \citep{li2020modification}. 

Recent advancements in artificial intelligence (AI) and machine learning (ML) offer new possibilities for improving predictive modeling in this area. AI-based models can analyze large experimental datasets to detect patterns and optimize design parameters, ultimately aiding in the development of more efficient anode structures \citep{farahani2024employing, coy2022bibliometric}. For instance, some studies have explored the use of ML and deep learning techniques to predict the performance of different anode materials and configurations, demonstrating how these approaches could reshape the future of MFC design \citep{panja2024enhancing,oyedeji2023optimal}. However, traditional analytical and empirical optimization approaches often struggle to capture the nonlinear and coupled dependencies between biological activity, material composition, and electrochemical performance in MFCs. As a result, identifying optimal operating conditions or structural configurations becomes computationally intensive and experimentally costly. Recent advances in AI-driven modeling have begun to overcome these challenges by learning complex mappings between process parameters and performance outcomes, enabling faster and more accurate prediction and optimization of MFC behavior. These developments set the stage for structure-aware machine learning approaches that can generalize across diverse anode geometries and materials. Data-driven optimization of this kind has enabled rapid identification of ideal operating conditions, providing valuable input for guiding more detailed 3D simulations \cite{SAYED20241015}.

Modeling complex 3D structures such as MFC anodes or other manufacturable designs requires capturing how geometry, material behavior, and functional performance interact in nonlinear ways \citep{10.1109/ssci51031.2022.10022022,10.1002/wcms.1711}. ML models have increasingly been used for this purpose and have revolutionized 3D structure prediction across engineering and materials science. Early data-driven models, including traditional Artificial Neural Networks (ANNs), provided faster approximations than physics-based simulations but struggled to encode geometry and typically relied on hand-crafted descriptors \citep{10.1039/d1sc01895g}. Physics-based simulations themselves remain accurate but slow and depend on simplifying assumptions that limit exploration of large design spaces \citep{10.1109/ssci51031.2022.10022022,10.1002/wcms.1711}.

To address the limitations of ANNs, deep learning architectures designed for geometry emerged. Graph Neural Networks (GNNs) became particularly effective in modeling spatial relationships by representing materials and their interactions as graph structures \citep{ganapathi2022graph}. While GNNs can capture complex inter-point relationships, they continue to face oversmoothing and scalability issues when applied to dense or highly detailed 3D geometries \citep{10.1109/ssci51031.2022.10022022,10.1002/wcms.1711,10.1038/s41467-021-23303-9, 10.48550/arxiv.2302.12177}.

Point-cloud architectures such as PointNet and PointNet++ were introduced to overcome some of these bottlenecks. PointNet directly processes raw 3D point-cloud data via global feature pooling, enabling efficient classification and segmentation without voxelization \citep{wang2019dynamic}. PointNet++ further incorporates hierarchical feature learning, allowing the model to better capture fine-grained local structures essential for complex geometric modeling tasks \citep{ruizhongtai2023deep}. However, despite these improvements, global pooling in point-cloud models can still lose fine structural cues critical for manufacturability or electrochemical predictions \citep{10.1088/1361-6501/ad6e0e,10.48550/arxiv.2112.09343}.

Overall, the existing methods for predicting MFC anode structures face several limitations, primarily related to their inability to effectively capture the complex geometries and interactions inherent in 3D designs \citep{kar2015category}. Traditional physics-based models often rely on simplifying assumptions that fail to capture the full complexity of material and geometric interactions, leading to imprecise predictions \citep{rajalingham2018large}. While GNNs excel at capturing relationships within large-scale graphs, they can be computationally demanding, limiting their use in real-time applications \citep{liu2023graph}. Similarly, while PointNet efficiently processes raw 3D data, its reliance on global pooling may cause a loss of important local geometric details \citep{schmidt2021crystal}.

To address these limitations, attention-based models have emerged as an alternative, dynamically prioritizing key features in the data to improve accuracy and robustness \citep{moradi2025single}. Specifically, a unified attention mechanism, which dynamically focuses on relevant features during training, enables models to filter out noise and prioritize useful information. This selective focus has significantly improved model generalization in complex and noisy environments \citep{yao2019parallel}. 
Yet, attention alone cannot fully capture the highly nonlinear dependencies between geometry and performance \citep{10.1109/wacv48630.2021.00357}. This gap has driven interest in the Kolmogorov–Arnold Network (KAN) \citep{liu2024kan}, a neural network architecture designed to approximate complex functions by hierarchically decomposing input data. KANs combine additive and multiplicative feature interactions based on Kolmogorov–Arnold representation theory, enabling them to capture highly nonlinear relationships in high-dimensional data \citep{he2023complex}. This capability, in conjunction with attention, improves the accuracy of AI-driven models in scientific applications, particularly in materials science and engineering, where nonlinear interactions play a critical role \citep{jang2021optimal}. In addition to these concepts, disentangled representation learning has become increasingly important in geometry-aware AI modeling. This approach allows models to separate different factors of variation within data, leading to more meaningful and interpretable feature representations \citep{wang2019dynamic}. Together, these techniques provide a more refined and structured approach to learning complex geometric patterns, making them particularly valuable for 3D structure prediction.

In summary, this paper presents \textbf{KANGURA}, a novel model for 3D structure prediction and provides the following primary contributions:

\begin{itemize}
    \item \textbf{KAN-based decomposition}, which improves function approximation and enhances the accuracy of complex 3D predictions.
    \item \textbf{Geometry-disentangled representation learning}, allowing the model to better capture structural variations by separating different geometric features.
    \item \textbf{Unified attention mechanisms}, which refine spatial awareness and help the model focus on the most relevant details for more reliable predictions.
    \item \textbf{Evaluation on both synthetic and real-world datasets}, demonstrating its effectiveness on the ModelNet40 benchmark and a real-world MFC anode structure with complex geometry.
\end{itemize}

\begin{table}[h]
    \centering
    \renewcommand{\arraystretch}{1.2} 
    \setlength{\tabcolsep}{5pt} 
    \small 
    
    \caption{Comparison of feature capabilities across different model types for 3D modeling}
    \label{tab:model_comparison1}
    
    \begin{adjustbox}{width=\textwidth} 
    \begin{tabular}{lcccccc}
        \toprule
        \textbf{Feature / Model Type} & 
        \makecell{\textbf{Geometric} \\ \textbf{Awareness}} & 
        \makecell{\textbf{Global Geometric} \\ \textbf{Understanding}} & 
        \makecell{\textbf{Local Geometric} \\ \textbf{Understanding}} & 
        \makecell{\textbf{Spatial} \\ \textbf{Attention}} & 
        \makecell{\textbf{Handles Complex} \\ \textbf{Functions}} & 
        \makecell{\textbf{Function} \\ \textbf{Decomposition}} \\
        \midrule
        Physics-Based Models \citep{moser2023modeling, mohan2020spatio,kang2024four,zhang2025physics}  & \xmark & \xmark & \xmark & \xmark & \xmark & \xmark \\
        Traditional ML Models \citep{huang2019traditional,hussin2020traditional,rojek2021traditional,nguyen2022ensemble} & \xmark & \xmark & \xmark & \xmark & \xmark & \xmark \\
        Deep Learning-Based Models \citep{zhou2018voxelnet,bello2020deep,xu2021learning,feng2019meshnet} & \cmark & \cmark & \cmark & \xmark & \cmark & \xmark \\
        Graph Neural Networks \citep{weng2020gnn3dmot,zou2021modulated,song2020meshgraphnet,wei2020view} & \cmark & \cmark & \cmark & \xmark & \cmark & \xmark \\
        PointNet Variants \citep{cao2022rp,zhao2022jsnet++,sheshappanavar2020novel,tian2024research} & \cmark & \cmark & \cmark & \xmark & \xmark & \xmark \\
        \textbf{KANGURA (Our Model)} & \cmark & \cmark & \cmark & \cmark & \cmark & \cmark \\
        \bottomrule
    \end{tabular}
    \end{adjustbox}
\end{table}

Table \ref{tab:model_comparison1} provides a comparative analysis of different model types based on their ability to capture geometric features and handle complex functions. Unlike traditional ML and physics-based models, deep learning-based approaches demonstrate improved geometric awareness. However, they lack function decomposition and spatial attention, which are fully addressed by \textbf{KANGURA}. Our model uniquely captures and refines the holistic and complementary 3D geometric semantics,  making it more effective for complex manufacturing designs.

The rest of the paper is organized as follows. The problem statement and dataset understanding are summarized in Section \ref{sec:Problem Setup}. Section \ref{sec:methodology} describes the preliminaries, definitions, and high-level steps of our proposed method. Section \ref{sec:results} highlights computational results and the comparative study. Finally, Section \ref{sec:conclusion} provides concluding remarks and suggestions
for future research.

\section{Problem Setup}
\label{sec:Problem Setup}
In this section, we define the problem of manufacturability classification in additive manufacturing. Specifically, our aim is to identify whether a given 3D design can be successfully manufactured, considering its geometric and structural attributes.  

\subsection{Problem Statement}
Additive manufacturing (AM) enables the creation of complex designs and functional assemblies, providing new possibilities for product development. A key difficulty, particularly for novice users of 3D fabrication technologies, is to ensure that their designs are compatible with the capabilities of the selected manufacturing process \citep{zhang2022web}. This challenge arises because of the intricate relationship between the design's geometry, material properties, and the manufacturing process capabilities. Manufacturability is influenced by various factors, including geometric complexity, structural feasibility, and material limitations. This challenge is particularly critical in applications such as microbial fuel cell (MFC) anodes, where intricate 3D structures are integral to performance. This research is important, not only because the design knowledge of manufacturers can be improved, but also because design exploration can be expanded and expedited. When feasible designs are generated faster, manufacturers can focus on designs with improved manufacturability, higher feasibility, and optimized performance. Addressing this issue through automated classification methods is essential to improving the design-to-production workflow, saving time, reducing costs, and enhancing design capabilities in various industries.

\begin{figure}[!htb]
		\begin{center}
			\fbox{\includegraphics[width=0.98\textwidth]{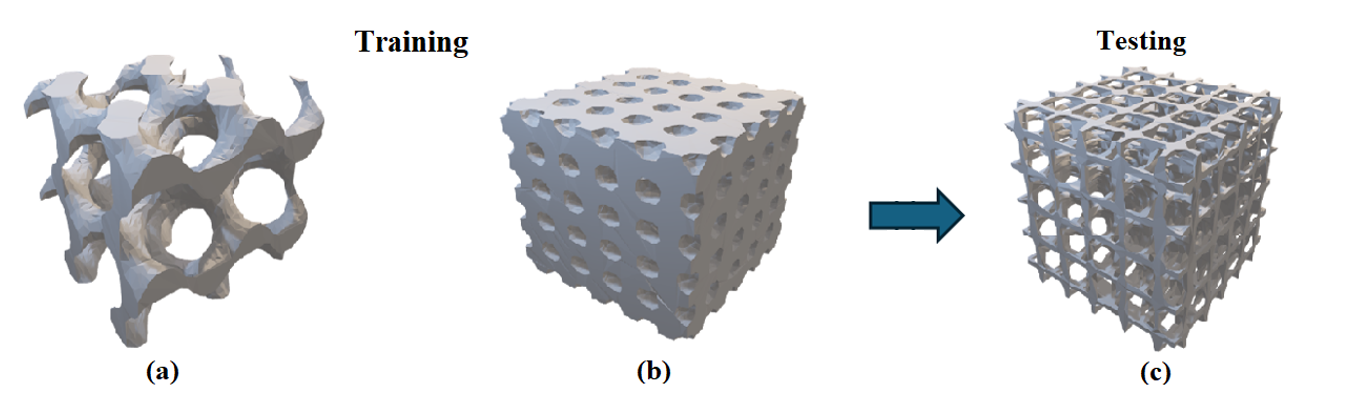}}
		\end{center}
		\caption{Illustration of the manufacturability classification process for 3D designs. Panel (a) shows manufacturable designs with feasible geometries. Panel (b) displays non-manufacturable designs with impractical features. Panel (c) demonstrates the testing phase, where the trained model is applied to predict the manufacturability of new designs.}
		\label{fig:2-1}
	\end{figure}

Figure \ref{fig:2-1} illustrates the problem setup for addressing manufacturability classification in the context of 3D designs. Panel (a) of Figure \ref{fig:2-1} shows the training phase, where manufacturable designs are represented by feasible geometries that can be physically fabricated. Panel (b) displays non-manufacturable designs, which are characterized by impractical features that restrict production. Finally, panel (c) represents the testing phase, where the trained model is applied to unseen 3D designs to predict their manufacturability. This workflow emphasizes the goal of automating the assessment process to accelerate validation and improve design outcomes.

\begin{figure}[!htb]
    \centering
    \fbox{\includegraphics[width=0.7\textwidth, height=0.4\textwidth, keepaspectratio]{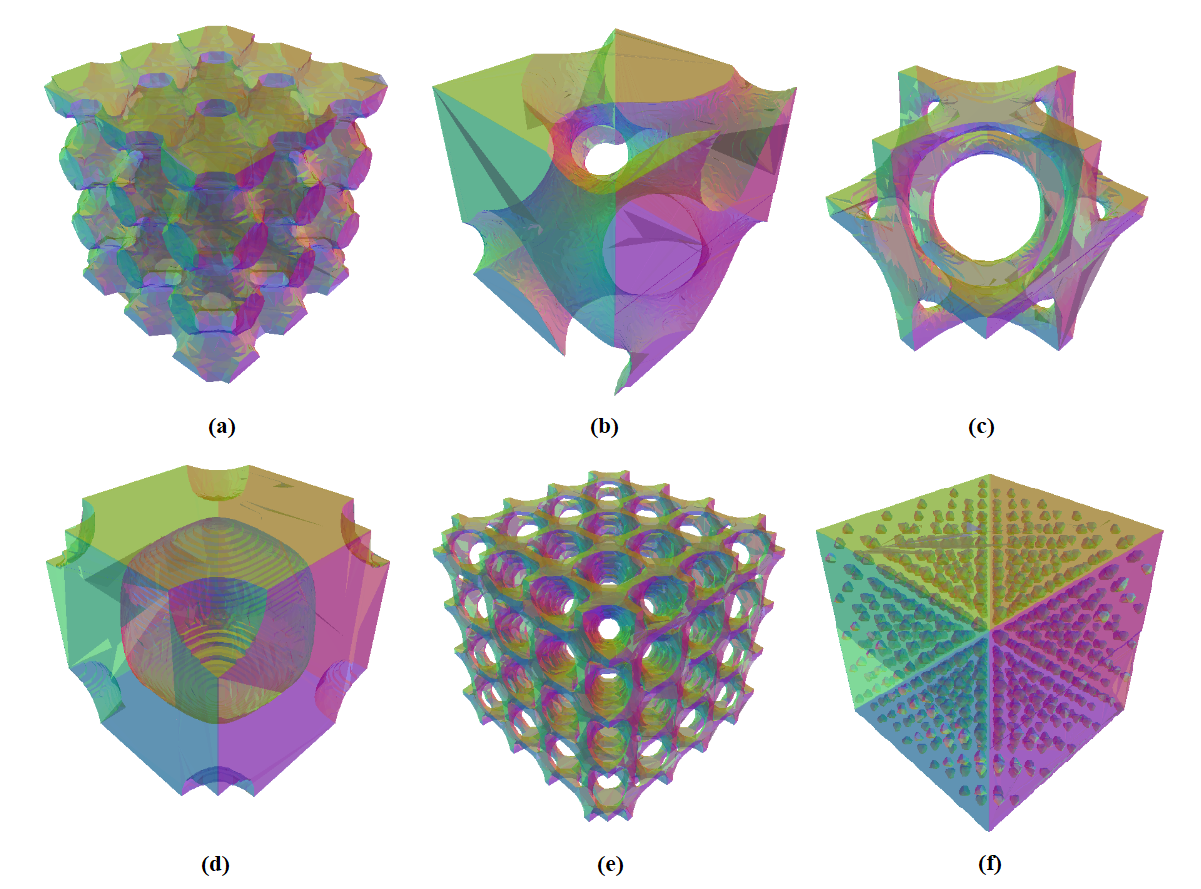}} 
    \vspace{0.1em} 
    \caption{3D visualizations of additional manufacturable and non-manufacturable designs, highlighting the internal structure of each object. Panels (a), (b), and (c) represent manufacturable designs, where the internal geometries are feasible for production. Panels (d), (e), and (f) display non-manufacturable designs, characterized by impractical internal features that prevent physical fabrication.}
    \label{fig:2-2}
\end{figure}
\subsection{Case Study}
To address this challenge, a dataset consisting of 1,000 3D models has been curated, including 300 manufacturable designs and 700 non-manufacturable designs. Each design is represented by its geometric and structural attributes, capturing the essential features that influence manufacturability. This class-imbalanced dataset reflects the reality of design data, where non-manufacturable configurations are more prevalent \citep{zeng2024data, zeng2025high, chilukuri2024generating}. The task is to develop a robust classification model capable of learning from these features and accurately distinguishing between manufacturable and non-manufacturable designs. The diversity in geometric complexity and structural variation within the dataset makes it a suitable benchmark for evaluating classification system performance. For instance, as shown in Figure \ref{fig:2-2}, manufacturable designs (panels (a), (b), and (c)) exhibit feasible internal geometries, while non-manufacturable designs (panels (d), (e), and (f)) display impractical internal features that complicate their fabrication.

\begin{figure}[!htb]
		\begin{center}
			\fbox{\includegraphics[width=0.7\textwidth]{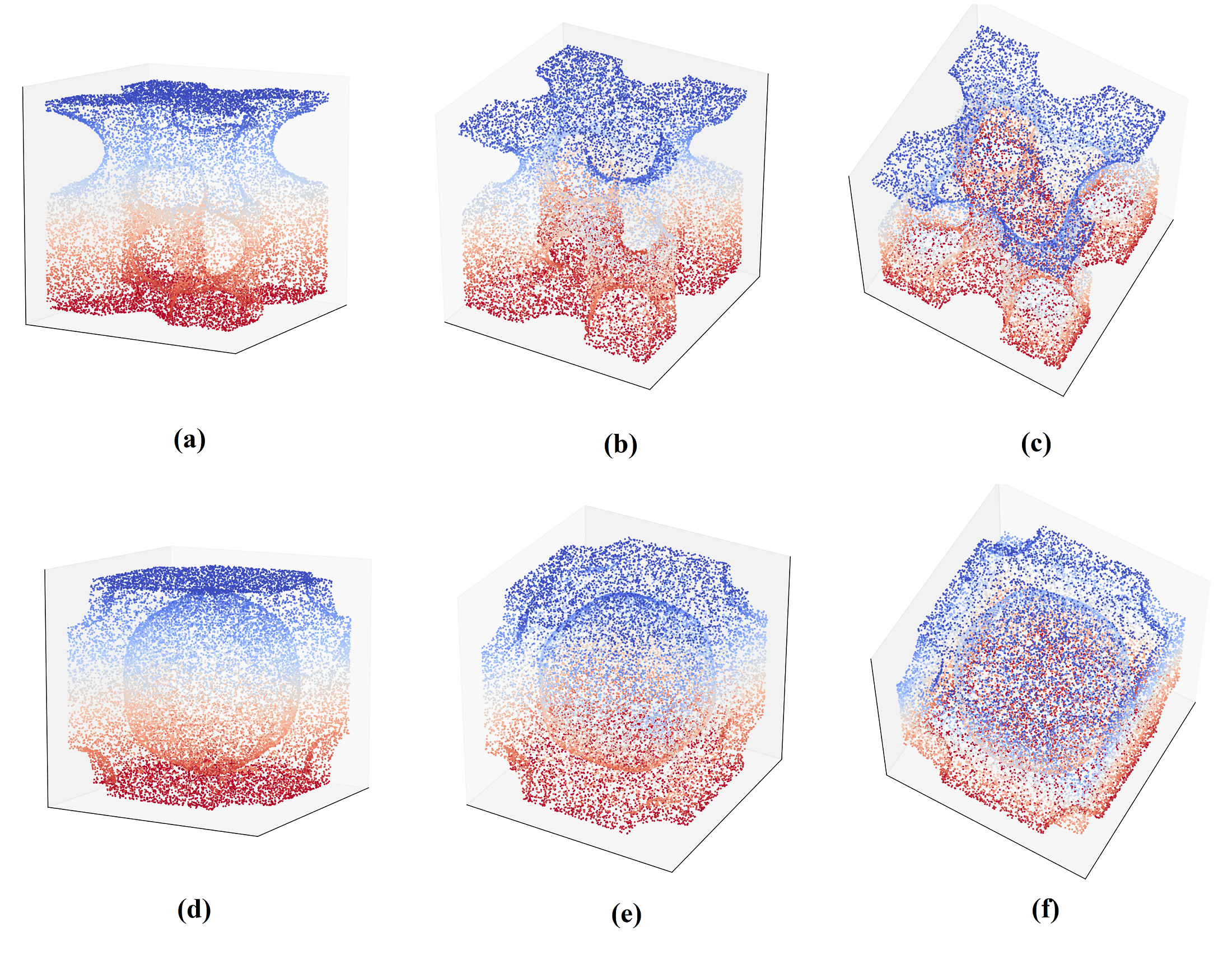}} 
		\end{center}
		\vspace{0.1em} 
		\caption{Multi-view point cloud visualizations of 3D designs used for manufacturability classification. Panels (a)–(c) show manufacturable designs with feasible internal geometries. Panels (d)–(f) illustrate non-manufacturable designs featuring impractical structural configurations that restrict fabrication.}
		\label{fig:2-3}
	\end{figure}

To effectively capture the spatial and structural properties of each design, point clouds are used as the primary 3D representation format in this study. Point clouds offer a high-fidelity and compact representation of object geometry, preserving fine-grained spatial details without requiring surface meshing or volumetric modeling. This enables the classification system to learn directly from raw geometric patterns, which are critical in assessing design feasibility. Their viewpoint-agnostic nature also allows consistent feature extraction across different orientations, enhancing the model's generalization capability. As illustrated in Figure \ref{fig:2-3}, point cloud renderings from multiple angles reveal the internal characteristics that influence manufacturability. Panels (a), (b), and (c) show different perspectives of manufacturable designs, each exhibiting coherent and production-ready internal structures. In contrast, panels (d), (e), and (f) depict non-manufacturable designs with irregular or impractical geometries that violate fabrication constraints.

\section{Methodology}
\label{sec:methodology}

\begin{figure*}[h]
\centering
\includegraphics[width=\textwidth]{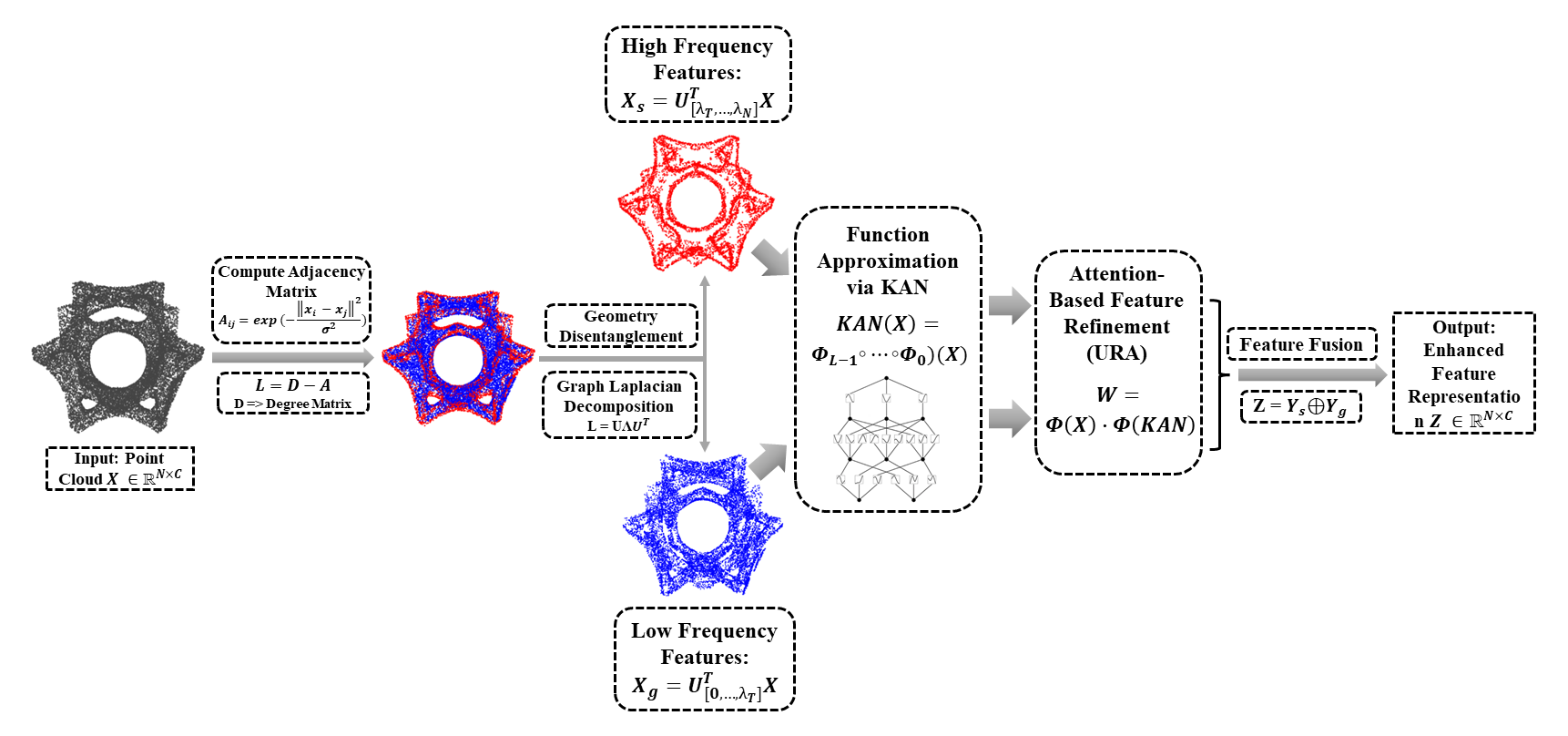} 
\caption{Flowchart of the KANGURA}
\label{fig:kangura_flowchart}
\end{figure*}
 
In this section, the high-level process of the proposed method is explained. The methodology shown in Algorithm \ref{alg:kangura} consists of five key steps: graph construction, geometry disentanglement, function approximation via KAN, attention-based feature refinement, and final feature fusion.

This framework is designed to learn a robust and enhanced geometric representation from 3D point cloud data, which can then be used for classification tasks such as determining manufacturability. The overall process, detailed below, transforms raw coordinates into a rich feature set that captures complex structural relationships.

\begin{algorithm}
    \caption{Overall Steps of \textbf{KANGURA}: \textbf{\underline{K}}olmogorov-\textbf{\underline{A}}rnold \textbf{\underline{N}}etwork-Based \textbf{\underline{G}}eometry-Aware Learning with \textbf{\underline{U}}nified \textbf{\underline{R}}epresentation \textbf{\underline{A}}ttention}
    \begin{algorithmic}[1]
        \State \textbf{Input:} Point cloud \(X \in \mathbb{R}^{N \times C}\) with \(N\) points and \(C\) features
        \State \textbf{Output:} Enhanced feature representation \(Z \in \mathbb{R}^{N \times C'}\)

        \textbf{Graph Construction and Spectral Representation}
        \State Compute adjacency matrix \( A_{ij} = \exp(-\|x_i - x_j\|^2 / \sigma^2) \) if \( \|x_i - x_j\| \leq \tau \), else \( 0 \). Compute graph Laplacian \( L = D - A \) and its decomposition \( L = U \Lambda U^T \). 
        
        \textbf{Geometry Disentanglement}
        \State Extract high-frequency sharp variation \( X_s = U_{[\lambda_T, \dots, \lambda_N]}^T X \) and low-frequency gentle variation \( X_g = U_{[0, \dots, \lambda_T]}^T X \). Apply spectral filtering: \( \hat{X}_s = (I - \tilde{A}) X_s \), \( \hat{X}_g = \tilde{A} X_g \).

        \textbf{KAN Functional Decomposition}
        \State Transform features using hierarchical KAN layers with functional mapping \( f(X) = \sum_{q=1}^{2n+1} \Phi_q \left( \sum_{p=1}^{n} \varphi_{qp}(X_p) \right) \). Compute layer-wise activations \( x_{l+1, j} = \sum_{i=1}^{n_l} \varphi_{l, j, i}(x_{l, i}) \). The final KAN representation is \( KAN(X) = (\Phi_{L-1} \circ \dots \circ \Phi_0)(X) \), applied separately to both sharp and gentle variations: \( KAN_s = KAN(\hat{X}_s) \), \( KAN_g = KAN(\hat{X}_g) \).

        \textbf{Unified Representation Attention}
        \State Compute attention weights \( W_s = \Theta_o(X) \cdot \Theta_s(KAN_s)^T \), \( W_g = \Phi_o(X) \cdot \Phi_g(KAN_g)^T \), then refine feature representations as \( Y_s = X + W_s (KAN_s) \), \( Y_g = X + W_g (KAN_g) \).

        \textbf{Feature Fusion and Output}
        \State Compute final representation \( Z = Y_s \oplus Y_g \) and return \( Z \).
    \end{algorithmic}
    \label{alg:kangura}
\end{algorithm}

\paragraph{Graph Construction and Spectral Representation}
To encode the geometric structure of a 3D point cloud \(X \in \mathbb{R}^{N \times C}\), we construct an adjacency matrix \(A \in \mathbb{R}^{N \times N}\) based on feature-space proximity:
\begin{equation}
    A_{ij} = 
    \begin{cases} 
        \exp\left(-\frac{\|x_i - x_j\|^2}{\sigma^2}\right), & \text{if } \|x_i - x_j\| \leq \tau \\
        0, & \text{otherwise}
    \end{cases}
\end{equation}
where \(\sigma\) is the scaling parameter and \(\tau\) is the distance threshold. This initial step transforms the raw point cloud into a structured graph, enabling the model to process intricate spatial relationships between points rather than treating them as an unordered set.

The graph Laplacian is computed as:
\begin{equation}
    L = D - A, \quad L = U \Lambda U^T
\end{equation}
where \(D\) is the degree matrix and \(U\) contains the eigenvectors. The computation of the graph Laplacian is critical as its spectral properties (eigenvectors and eigenvalues) form the basis for frequency-based filtering and analysis of the underlying geometry in subsequent steps.

\paragraph{Geometry Disentanglement}
To separate fine-grained and coarse geometric variations, we partition the spectral domain into sharp and gentle components. High-frequency components (sharp variations) are extracted as:
\begin{equation}
    X_s = U_{[\lambda_T, \dots, \lambda_N]}^T X, 
\end{equation}
while low-frequency components (smooth variations) are given by:
\begin{equation}
    X_g = U_{[0, \dots, \lambda_T]}^T X.
\end{equation}
These components are further refined via:
\begin{equation}
    \hat{X}_s = (I - \tilde{A}) X_s, \quad \hat{X}_g = \tilde{A} X_g,
\end{equation}
ensuring structured feature disentanglement.

This key step leverages the spectral representation to partition the geometry into more meaningful and interpretable components. The high-frequency "sharp" variations correspond to fine-grained local details like edges and corners, while the low-frequency "gentle" variations represent the coarse, overall shape of the object. By separating these factors of variation, the model can learn from both types of geometric information independently, ensuring that subtle but important details are not overshadowed by the global structure.

\paragraph{KAN for Functional Decomposition}
Instead of traditional MLPs, KANGURA employs KAN for function approximation. The Kolmogorov-Arnold representation theorem states that any multivariate function \(f: \mathbb{R}^n \to \mathbb{R}\) can be represented as:
\begin{equation}
    f(X) = \sum_{q=1}^{2n+1} \Phi_q \left( \sum_{p=1}^{n} \varphi_{qp}(X_p) \right),
\end{equation}
where \(\varphi_{qp}\) and \(\Phi_q\) are learnable univariate functions. The KAN is applied hierarchically as \( x_{l+1, j} = \sum_{i=1}^{n_l} \varphi_{l, j, i}(x_{l, i}),\) allowing adaptive functional decomposition. 
According to mathematical theory, KANs can capture complex, non-linear relationships in high-dimensional data more effectively than conventional models. This is particularly valuable for modeling 3D structures where the interplay between geometric features is inherently non-linear.
The overall KAN representation is:
\begin{equation}
    \text{KAN}(X) = (\Phi_{L-1} \circ \Phi_{L-2} \circ \dots \circ \Phi_0)(X),
\end{equation}
which enhances feature extraction from the disentangled sharp and gentle variations:
\begin{equation}
    KAN_s = \text{KAN}(\hat{X}_s), \quad KAN_g = \text{KAN}(\hat{X}_g).
\end{equation}

\paragraph{Unified Representation Attention  (URA) for Feature Refinement}
To refine features, KANGURA employs Unified Representation Attention by computing attention weights via bilinear transformations:
\begin{equation}
    W_s = \Theta_o(X) \cdot \Theta_s(KAN_s)^T, \quad W_g = \Phi_o(X) \cdot \Phi_g(KAN_g)^T.
\end{equation}
The feature refinement follows \(Y_s = X + W_s KAN_s\) , and \(Y_g = X + W_g KAN_g\). URA dynamically balances local and global feature aggregation, enhancing feature expressiveness.

\paragraph{Feature Fusion and Final Output}
The final representation integrates sharp and smooth variations:
\begin{equation}
    Z = Y_s \oplus Y_g.
\end{equation}
This fusion step ensures that both fine-grained local detail and global structural coherence are preserved. This combined representation is then returned as the final output of the model. By ensuring that the holistic and complementary 3D geometric semantics are captured, this final vector is made highly effective for complex classification tasks, such as those in advanced manufacturing. As illustrated in Figure~\ref{fig:kangura_flowchart}, the KANGURA framework handles point cloud data across five essential phases: Graph Construction, Geometry Disentanglement, Function Approximation via KAN, Attention-Based Feature Refinement, and Feature Fusion. The procedure commences with the creation of a graph from the point cloud, utilizing an adjacency matrix and graph Laplacian. Following this, geometry disentanglement is performed, which isolates the high and low-frequency elements of the point cloud's geometry to seize fine-grained geometric details.

In the subsequent stages, KANGURA utilizes Function Approximation via KAN to project the point cloud features into a target output space. These features are then enhanced through an attention-based mechanism that amplifies the significant features. Ultimately, the high and low-frequency features are combined to produce a refined feature representation for use in subsequent tasks.

\section{Results}
\label{sec:results}
This section presents the evaluation of KANGURA on both benchmark and real-world datasets. The first subsection compares its performance against state-of-the-art (SOTA) models on a standard 3D classification benchmark, while the second assesses its effectiveness in the real-world MFC anode structure classification.

\begin{figure}[h]
    \centering
    \begin{minipage}{0.50\textwidth} 
        \centering
        \small
        \renewcommand{\arraystretch}{1.1}
        \setlength{\tabcolsep}{4pt}
        \captionof{table}{Comparison of different models on ModelNet40 benchmark dataset}
        \label{tab:model_comparison}
        \begin{adjustbox}{width=\textwidth}
        \begin{tabular}{lccc}
            \toprule
            \textbf{Method} & \textbf{Input Type} & \textbf{Accuracy (\%)} & \textbf{$\Delta$  from Kangura} \\
            \midrule
            SPH & Mesh & 68.2 & 24.5 \\
            LFD & View & 75.5 & 17.2 \\
            FPNN & Volume & 88.4 & 4.3 \\
            PointNet & 1K points & 89.2 & 3.5 \\
            SCN & 1K points & 90.0 & 2.7 \\
            MVCNN & View & 90.1 & 2.6 \\
            Pairwise & View & 90.7 & 2.0 \\
            PointNet++ & 1K points & 90.7 & 2.0 \\
            KCNet & 1K points & 91.0 & 1.7 \\
            Kd-Net & Point & 91.8 & 0.9 \\
            MeshNet & Mesh & 91.9 & 0.8 \\
            PointNet++ & 5K points + normal & 91.9 & 0.8 \\
            3D-GCN & 1K points & 92.1 & 0.6 \\
            PointCNN & 1K points & 92.2 & 0.5 \\
            PointWeb & 1K points & 92.3 & 0.4 \\
            PointConv & 1K points + normal & 92.5 & 0.2 \\
            FPConv & 1K points & 92.5 & 0.2 \\
            Point2Sequence & 1K points & 92.6 & 0.1 \\
            \textbf{KANGURA (Ours)} & Point & \textbf{92.7} & \textbf{0.0} \\
            \bottomrule
        \end{tabular}
        \end{adjustbox}
    \end{minipage}
\end{figure}

\begin{figure}[ht]
    \centering
    \begin{minipage}{0.60\textwidth} 
        \centering
        \includegraphics[width=\textwidth]{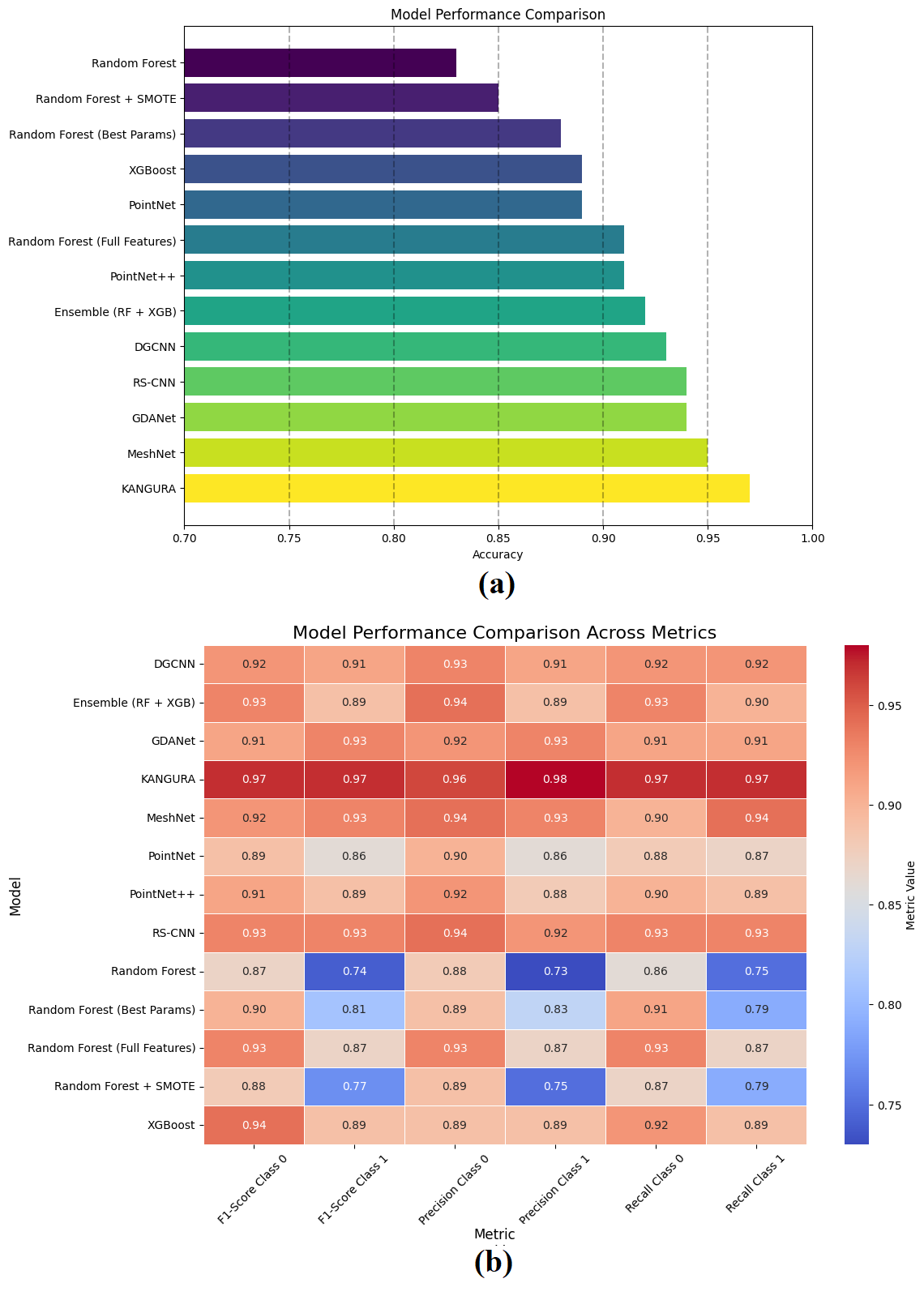} 
    \end{minipage}
    \caption{Performance analysis on the real-world data: (a) Accuracy comparison, (b) Heatmap of models performance across Precision, Recall, and F1-Score} 
    \label{fig:my_figure}
\end{figure}

\subsection{Benchmark Performance}

To assess the effectiveness of \textbf{KANGURA}, we compare its classification accuracy against over 15 state-of-the-art models on the ModelNet40 benchmark \citep{wu20153d}, a widely adopted dataset for 3D object recognition comprising 40 shape categories. As presented in Table~\ref{tab:model_comparison}, \textbf{KANGURA achieves an accuracy of 92.7\%}, surpassing the performance of leading methods evaluated.

Mesh-based approaches such as SPH (68.2\%) and MeshNet (91.9\%) illustrate the potential of utilizing explicit surface representations; however, their performance remains below that of leading point-based models, suggesting limitations in capturing fine-grained local structures. View-based methods, including MVCNN (90.1\%) and Pairwise (90.7\%), aggregate visual information across multiple projections but are inherently constrained by discretized viewpoints and lack of spatial continuity. Volumetric techniques such as FPNN (88.4\%) are are additionally limited by resolution degradation and the high computational cost introduced by voxelization.

In contrast, point-based models consistently yield higher performance, with top contenders including Point2Sequence (92.6\%), FPConv (92.5\%), and PointConv (92.5\%). \textbf{KANGURA} surpasses each of these baselines, with relative improvements of 0.1\%, 0.2\%, and 0.2\%, respectively. These results underscore the effectiveness of KANGURA’s architectural innovations, the KAN-based functional decomposition and the unified attention mechanism, which jointly contribute to superior geometric feature learning. The resulting enhancement in classification accuracy highlights the model’s robustness and generalizability across diverse input modalities.

The accuracy gains are particularly significant given that earlier models predominantly rely on point-based convolutional networks or graph-based architectures. By introducing a structured function decomposition, \textbf{KANGURA outperforms SOTA models} by refining feature extraction and improving spatial awareness. This enables a more robust and precise understanding of 3D structures, ultimately setting a new benchmark in 3D shape recognition. The delta accuracy column strengthens this claim, as it highlights \textbf{KANGURA}'s superior ability to generalize and handle variations in input data compared to previous methods.

\subsection{Real-world Data}

To evaluate KANGURA's real-world applicability, we tested its performance on the MFC anode structure classification, comparing it with 15 different models, including traditional machine learning models using numerical features (e.g., topological and surface ratio features) and deep learning models utilizing 3D mesh data.

Figure~\ref{fig:my_figure}(a) shows that \textbf{KANGURA achieves the highest accuracy (97\%)}, surpassing MeshNet (95\%) and GDANet (94\%), as well as vanilla machine learning models. While ML models capture key geometric properties, they lack the ability to represent local-global spatial dependencies within 3D structures. MeshNet, trained on mesh-based representations, outperforms numerical feature-based approaches but remains constrained by standard convolutions, which limits its ability to fully capture hierarchical geometric relationships. Furthermore, as shown in Figure~\ref{fig:my_figure}(b), KANGURA maintains superior precision, recall, and F1 score in both feasible (Class 1) and infeasible (Class 0) designs, ensuring a robust classification essential to optimize the 3D structure in MFC applications.

\section{Conclusion}
\label{sec:conclusion}
In this study, we introduced KANGURA, a geometry aware Kolmogorov Arnold Network designed to overcome the core limitations of existing three dimensional machine learning approaches in advanced manufacturing. Our work demonstrates that KANGURA can successfully model the intricate relationships between complex geometry, material behavior, and functional performance. Although the MFC anode classification task served as our primary example, the architecture is broadly applicable to quality control, reliability engineering, materials design, and structural optimization. Through extensive evaluation, we compared KANGURA against 18 SOTA models on a benchmark dataset and 12 models on real-world data, achieving the highest accuracy across both settings. KANGURA surpasses leading deep learning models, as well as traditional machine learning approaches using geometric and topological features.

The performance improvements arise because the individual components of the architecture work both independently and in combination to capture features that other models fail to represent. The Kolmogorov Arnold functional decomposition enables the model to describe complex nonlinear interactions in geometric data. The geometry disentangled representation learning mechanism separates global structure from localized features that are crucial for manufacturing quality. The unified attention mechanism focuses the model on structurally informative regions while preserving global coherence. When combined, these components allow KANGURA to learn multiscale geometric dependencies that are essential for high fidelity prediction. These results highlight its potential for data-driven optimization in advanced manufacturing applications, where precise 3D modeling is critical for process efficiency and quality assurance. Future work will explore broader industrial applications, further enhancing its interpretability and scalability across various material science and engineering domains.


\section*{Acknowledgements}

We would like to express our profound appreciation to the Department of Industrial \& Management Systems Engineering (IMSE), WVU for their consistent and invaluable support during this research project. We are also immensely thankful to our colleagues and experts in the area of smart manufacturing for thier valuable support.

\medskip
\small
\medskip
\bibliographystyle{unsrtnat}
\bibliography{references}

\end{document}